\documentclass[PHME, 2026]{PHMSociety}

\usepackage{graphicx}
\usepackage{placeins}
\usepackage{float}
\usepackage{amsmath}
\usepackage{amssymb}
\usepackage{booktabs}
\usepackage{colortbl}
\usepackage{xurl}

\definecolor{rowgray}{gray}{0.93}
\graphicspath{{Figures/}} 

\begin{document}

\title{Virtual Temperature Sensors in Power Transformers Using Neural Ordinary Differential Equations}

\author{%
	Berk Hadzhamolla\authorNumber{1}, Alexander Johannes Stasik\authorNumber{2}, and Signe Riemer-Sørensen\authorNumber{3}
}

\address{
	\affiliation{{1}}{University of Oslo, Oslo, 0316, Norway}{ 
		{\email{berkh@uio.no}}
		} 
	\affiliation{2,3}{SINTEF AS, Department of Mathematics and Cybernetics, Oslo, Norway}{ 
		\email{signe.riemer-sorensen@sintef.no} \\
        \email{alexander.stasik@sintef.no}
		} 
        \affiliation{2}{Department of Data Science, Norwegian University of Life Sciences, Ås, Norway}{ 
		{\email{alexander.johannes.stasik@nmbu.no}}		} 
}

\maketitle
\pagestyle{fancy}
\thispagestyle{plain}

\phmLicenseFootnote{Berk Hadzhamolla}

\begin{abstract}
Accurate modeling and forecasting of power transformer thermal behavior are critical for ensuring reliability, extending asset lifetime, and enabling optimized power system operation. Numerical approaches, such as finite element methods (FEM) and computational fluid dynamics (CFD), offer high fidelity but suffer from prohibitive computational costs, complex mesh generation, and limited feasibility in real-time or large-scale applications, as well as often unknown geometries. Lumped-parameter thermal models provide a more practical alternative but depend on transformer-specific thermal constants and often fail to capture dynamic responses under varying operating and environmental conditions. Purely data-driven machine learning (ML) methods, including artificial neural networks (ANNs), convolutional neural networks (CNNs), and recurrent architectures such as long short-term memory (LSTM) networks, have shown success in forecasting transformer oil, winding, and hotspot temperatures; however, they typically require large volumes of high-quality training data and risk producing physically inconsistent or uninterpretable results. To overcome these limitations, hybrid frameworks such as physics-informed neural networks (PINNs) embed physical laws into the learning process, enabling physically consistent solutions while reducing data demands. This paper applies a physics-aware modeling of Neural Ordinary Differential Equations (Neural ODEs) adapted for forecasting transformer thermal behavior using real-world time-series data. Neural ODEs model system dynamics in continuous time, enabling smoother predictions, robustness to irregular sampling, and improved extrapolation capabilities compared to discrete-time models such as LSTMs. A key contribution of this work is the integration of simplified heat-transfer equations for power transformers directly into the Neural ODE, enabling a physics-aware formulation of the thermal dynamics. The model’s performance and generalization capabilities are evaluated across datasets from fifteen distinct transformers located in different regions of Norway, and characterized by varying designs and cooling mechanisms. The results demonstrate the success of the developed Neural ODEs framework to serve as a standardized, physics-aware, and robust forecasting tool for heterogeneous transformer units.
\end{abstract}

\section{Introduction}
\label{sec:introduction}

Power transformers are essential components of modern electrical grids, enabling efficient transmission and distribution of electrical energy through voltage conversion. Their continuous operation is subjected to fluctuating load demands and varying environmental conditions, which introduce significant thermal stress. Excessive operating temperatures accelerate insulation ageing processes, reduce transformer service life, and increase the risk of costly failures or large-scale outages. Consequently, accurate estimation and forecasting of transformer internal temperatures (virtual sensing) has become a critical requirement for condition monitoring, overload control, reliable grid operation and predictive maintenance strategies \cite{Swift2001,Lundgaard2004,Susa2005,Piercy1994}.

Among the most important thermal indicators of transformer health are the top-oil and winding hot-spot temperatures, as these quantities reflect cooling performance and directly influence the ageing rate of cellulose insulation \cite{Lundgaard2004}. Accurate estimation of these thermal states is therefore essential for transformer lifetime assessment and operational reliability. To enable effective online thermal condition monitoring, dynamic thermal models capable of predicting internal transformer temperatures under varying operating and environmental conditions are required \cite{Swift2001,Susa2005}.

Prevalent methods for estimating transformer internal temperatures rely on physics-based models or empirical approaches. Traditionally, detailed three-dimensional (3D) models based on FEM and CFD have been employed to solve coupled heat transfer and fluid flow equations governing transformer thermal behaviour \cite{Yan2025LSTM,JuarezBalderas2020HotSpot}. Such numerical approaches provide detailed representations of transformer geometry and internal oil circulation and were originally developed as high-fidelity alternatives to semi empirical analytical methods based on experimentally derived thermal constants \cite{Oommen2009TransformerDesign}. These models aim to capture spatial temperature distributions and internal flow dynamics with high physical fidelity. However, their practical application is often limited by high computational cost, sensitivity to boundary conditions, and incomplete knowledge of internal physical parameters.

An alternative involves lumped-parameter thermal circuit models, which approximate heat transfer dynamics using equivalent thermal resistances and capacitances \cite{Bragone2022PINNsThermal}. While computationally efficient, these models struggle to accurately capture dynamic thermal behaviour under highly variable loads or environmental conditions. Efforts to improve their accuracy have included incorporating temperature-dependent losses, oil viscosity, and external factors such as solar radiation \cite{Yan2025LSTM}.

An important aspect of transformer thermal behaviour is the cooling system, which dissipates heat generated by core and winding losses. Cooling configurations such as ONAN (Oil Natural, Air Natural), ONAF (Oil Natural, Air Forced), OFAF (Oil Forced, Air Forced), and ODAF (Oil Forced, Water Forced) significantly influence thermal performance and loading capability \cite{williams2024cooling}. This diversity introduces substantial variability, making the development of generalized thermal models particularly challenging.

The increasing deployment of high-resolution sensors in modern transformers has enabled data-driven modelling approaches based on ML. Early studies using ANNs focused on top-oil temperature estimation and later expanded to winding and hot-spot forecasting \cite{JuarezBalderas2020HotSpot,Wei2017HotspotForecasting,Yan2025LSTM,Vilaithong2007NeuralNetTopOil}. ANNs demonstrated strong predictive capability and, in several studies, improved accuracy compared to semi-physical models under rapidly changing ambient conditions \cite{Vilaithong2007NeuralNetTopOil,Temboa2022DataDriven}. Their primary advantage lies in learning nonlinear mappings directly from data without requiring explicit physical formulations \cite{JuarezBalderas2020HotSpot} and thereby bypassing a weakness of the physical models.

With the rise of deep learning, more advanced architectures such as Long Short-Term Memory (LSTM) networks and Convolutional Neural Networks (CNNs) have been explored, proving effective in capturing sequential dependencies and spatiotemporal features in operational data \cite{Tan2019UltraShortTerm,Rubasinghe2023CNNLSTM,Yan2025LSTM}. Recently, models such as Time-series Dense Encoder (TiDE) and Temporal Convolutional Networks (TCN) have also been applied to transformer temperature forecasting, demonstrating improvements over IEC~60076-7 benchmark approaches \cite{Temboa2022DataDriven}. Despite these advances, purely data-driven methods often require large datasets and may lack physical interpretability or robust generalization across different transformer designs and cooling systems \cite{Huang2022PINNsReview,Karniadakis2021,Bragone2022PINNsThermal}.

Hybrid and physics-informed approaches have emerged to address these limitations. Bragone et al.~\cite{Bragone2022PINNsThermal} applied Physics-Informed Neural Networks (PINNs) using a one-dimensional heat diffusion equation to model transformer thermal dynamics, achieving high accuracy compared to finite-volume simulations. More broadly, PINNs have been successfully applied in power-system applications including state estimation, dynamic analysis, and optimal power flow \cite{Huang2022PINNsReview,Gao2023PhysicsGCN}. Through specially designed loss functions, PINNs encourage physically consistent solutions and thereby improve interpretability and improve performance in data-scarce scenarios.

In parallel, Neural Ordinary Differential Equations (Neural ODEs) have emerged as another promising hybrid approach. Neural ODEs were introduced by Chen et al.~\cite{Chen2018NeuralODE}, where neural networks parameterize continuous-time system dynamics integrated through numerical ODE solvers. As discussed by Kidger~\cite{Kidger2021NeuralDE}, this framework combines the flexibility of deep learning with the structure of numerical integration, providing an effective representation for scientific machine learning tasks. Neural ODEs can be interpreted as continuous-depth generalizations of Residual Networks (ResNets) \cite{He2016DeepResidual,Chen2018NeuralODE}, often yielding smoother trajectories and improved stability in long-horizon forecasting problems. Continuous-time modeling is attractive for PHM systems because sensor data are often irregular, missing, or asynchronous.

This work applies a physics-aware Neural ODE framework to forecast transformer thermal behaviour using real-world operational data from fifteen transformers across Norway. The remainder of this paper is organized as follows. Section~\ref{sec:methodology} describes the proposed methodology, including the transformer thermal model and the Neural ODE formulation. Section~\ref{sec:neural_ode} presents the physics-aware Neural ODE framework. Section~\ref{sec:results} reports and discusses the results. Section~\ref{sec:conclusion} concludes the paper.


\section{Methodology}
\label{sec:methodology}

\subsection{Transformer Thermal Dynamics}
\label{subsec:thermal_dynamics}

The thermal behavior of power transformers is governed by coupled heat transfer
between windings, cooling oil, and the ambient
environment~\cite{Swift2001,Susa2005}. Treating the windings as the primary heat
source and the oil as an intermediate thermal buffer, the coupled energy balance
reads:

\begin{align}
  C_{p,w} \frac{d T_{\mathrm{w}}}{dt} &= -h_1 \bigl(T_{\mathrm{w}} -
    T_{\mathrm{oil}}\bigr) + L_E(P), \label{eq:winding_ode} \\[4pt]
  C_{p,o} \frac{d T_{\mathrm{oil}}}{dt} &= h_1 \bigl(T_{\mathrm{w}} -
    T_{\mathrm{oil}}\bigr) - h_2 \bigl(T_{\mathrm{oil}} -
    T_{\mathrm{amb}}\bigr), \label{eq:oil_ode}
\end{align}

\noindent where $C_{p,w}$ and $C_{p,o}$ are the thermal capacitances of the
winding and oil, $h_1$ is the winding-to-oil heat transfer coefficient, $h_2$
is the oil-to-ambient heat transfer coefficient, and $L_E(P)$ represents
electrical losses as a function of active power load $P$.

In practice, the thermal constants $C_{p,w}$, $C_{p,o}$, $h_1$, and $h_2$ are
transformer-specific, often unknown, and vary with cooling configuration and
operating conditions. Rather than identifying these parameters explicitly, the
internal thermal states are collected into a state vector $\mathbf{y}(t)$ and
the external drivers into a control vector $\mathbf{x}(t)$:

\begin{equation}
  \mathbf{y}(t) =
  \begin{bmatrix}
    T_{\mathrm{oil}}(t) \\
    T_{\mathrm{hw}}(t) \\
    T_{\mathrm{lw}}(t) \\
    T_{\mathrm{hhs}}(t) \\
    T_{\mathrm{lhs}}(t)
  \end{bmatrix} \in \mathbb{R}^5,
  \qquad
  \mathbf{x}(t) =
  \begin{bmatrix}
    P_{\mathrm{hv}}(t) \\
    P_{\mathrm{lv}}(t) \\
    T_{\mathrm{amb}}(t)
  \end{bmatrix} \in \mathbb{R}^3.
  \label{eq:state_control}
\end{equation}

The state vector $\mathbf{y}(t)$ contains oil temperature, high- and low voltage
winding temperatures, and high- and low voltage hotspot temperatures. The control
vector $\mathbf{x}(t)$ contains high- and low-voltage active power loads and
ambient temperature. For transformers with missing measurements, the corresponding
entries are excluded from $\mathbf{y}$ or $\mathbf{x}$ accordingly.

Equations~\eqref{eq:winding_ode}--\eqref{eq:oil_ode} reveal two structural
properties that directly inform the proposed architecture. First, the rate of
change of each thermal state depends only on the current states $\mathbf{y}(t)$
and current external inputs $\mathbf{x}(t)$, with no explicit dependence on
past history, yielding a first-order Markovian state-space structure. Second,
$\mathbf{y}(t)$ and $\mathbf{x}(t)$ play fundamentally different physical
roles: $\mathbf{y}(t)$ represents the internal thermal states of the system,
while $\mathbf{x}(t)$ comprises the external drivers of heat generation and
dissipation. The proposed physics-aware Neural ODE preserves both properties
explicitly in its architecture, as described in Section~\ref{sec:neural_ode}.

The thermal dynamics are therefore expressed as the general data-driven
formulation:

\begin{equation}
  \frac{d\,\mathbf{y}(t)}{dt} = F\!\bigl(\mathbf{y}(t),\,\mathbf{x}(t);\,\theta\bigr),
  \label{eq:general_ode}
\end{equation}

\noindent where $F$ is a learned function parameterized by $\theta$.


\section{Physics-Aware Neural ODE Framework}
\label{sec:neural_ode}

\subsection{Neural ODEs with Exogenous Forcing}
\label{subsec:neural_ode_formulation}

Purely data-driven models such as LSTMs predict $\mathbf{y}(t+1)$ directly
from a window of past observations, with no structural constraint enforcing
the physical relationships in
Equations~\eqref{eq:winding_ode}--\eqref{eq:oil_ode}. Over long autoregressive
rollouts, this allows trajectories to drift in physically implausible directions.
To address this, the proposed physics-aware Neural ODE replaces the right-hand
side of Equation~\eqref{eq:general_ode} with a neural network
$\mathcal{NN}_\theta$, so that the state trajectory is obtained by numerical
integration of a learned vector field:

\begin{equation}
  \mathbf{y}(t) = \mathbf{y}(t_0) + \int_{t_0}^{t}
    \mathcal{NN}_\theta\!\bigl(\mathbf{y}(t'),\,\mathbf{x}(t')\bigr)\,dt'.
  \label{eq:neural_ode_integral}
\end{equation}

The physics-awareness of the framework is embedded into the architecture in
two concrete ways, directly mirroring the structure of
Equations~\eqref{eq:winding_ode}--\eqref{eq:oil_ode}. First, the inputs to
$\mathcal{NN}_\theta$ at each integration step are exactly
$(\mathbf{y}(t), \mathbf{x}(t))$, preserving the physical decomposition
between internal thermal states and external excitations. Second, the network
always outputs $d\mathbf{y}/dt$ rather than predicting temperatures directly,
enforcing the continuous time heat-transfer structure at every integration step.

The exogenous input $\mathbf{x}(t')$ is required at every integration step.
Since $\mathbf{x}$ is observed only at discrete times $\{t_0, t_1, \ldots,t_N\}$ with interval $\Delta t = 900\,\mathrm{s}$, continuous-time
access is provided by piecewise-linear interpolation, where $k$ denotes the index of the most recent observation such that $t_k \leq t < t_{k+1}$:

\begin{equation}
  \mathbf{x}(t) = \mathbf{x}_{k} + \frac{t - t_k}{\Delta t}
    \bigl(\mathbf{x}_{k+1} - \mathbf{x}_{k}\bigr),
  \qquad t \in [t_k,\, t_{k+1}).
  \label{eq:interpolation}
\end{equation}

This construction of a neural network driven by a continuously interpolated
control path is closely related to Neural Controlled Differential Equations
(Neural CDEs)~\cite{Kidger2020NeuralCDE}, which model the influence of external
signals on system dynamics as $d\mathbf{y}(t) = f_\theta(\mathbf{y}(t))\,dX(t)$,
where $X(t)$ is the continuously interpolated control path. In the present
formulation, $\mathbf{x}(t)$ enters the network directly as a concatenated
input rather than through a channel matrix, yielding a computationally simpler
but structurally equivalent treatment of the exogenous driving signal.

\subsection{Real life scenario testing}
The NeuralODE is implemented and tested on 15 transformers in the Norwegian transmission grid\footnote{Data provided by Statnett.}. Each dataset spans multiple months or years of operational data formatted as a multivariate time series capturing a range of thermal and electrical measurements recorded every 900 seconds (15 minutes) as outlined in Equation~\ref{eq:state_control}.Minor gaps, defined as missing intervals shorter than four consecutive samples (1 hour), were filled using linear interpolation, preserving temporal continuity without distorting the underlying time-series patterns. Longer gaps were excluded from the analysis to avoid introducing artificial thermal dynamics and interpolation bias.

\subsection{Model Architecture and Implementation}
\label{subsec:architecture}

The vector field $\mathcal{NN}_\theta$ is modelled with a three-layer fully connected network
with hidden dimension $H = 256$ and $\tanh$ activations. Let
$\mathbf{z} = [\mathbf{y};\mathbf{x}] \in \mathbb{R}^{n_y + n_x}$ denote the
concatenated input at each integration step. The forward pass is:

\begin{align}
  \mathbf{h}_1 &= \tanh(W_1 \mathbf{z} + \mathbf{b}_1), \label{eq:h1} \\
  \mathbf{h}_2 &= \tanh(W_2 \mathbf{h}_1 + \mathbf{b}_2), \label{eq:h2} \\
  \mathcal{NN}_\theta(\mathbf{y},\mathbf{x}) &= W_3 \mathbf{h}_2 + \mathbf{b}_3. \label{eq:output}   
\end{align}

\noindent where $W_1 \in \mathbb{R}^{H \times (n_y+n_x)}$,
$W_2 \in \mathbb{R}^{H \times H}$, $W_3 \in \mathbb{R}^{n_y \times H}$,
and $\mathbf{b}_1, \mathbf{b}_2, \mathbf{b}_3$ are the corresponding bias
vectors forming the trainable parameters $\theta$. Gradients with respect to $\theta$ are computed by backpropagation through the adaptive integration steps of the \texttt{dopri5} solver via standard automatic differentiation, as implemented in \texttt{torchdiffeq}~\cite{Chen2018NeuralODE}.

\subsection{Training Procedure and Hyper Parameters}
\label{subsec:training}


The network parameters $\theta$ are optimized by minimizing a composite loss:

\begin{equation}
  \mathcal{L}(\theta) = \mathcal{L}_{\mathrm{MSE}}(\theta) +
    \lambda\,\mathcal{L}_{\mathrm{smooth}}(\theta),
  \label{eq:loss}
\end{equation}

\noindent where

\begin{align}
  \mathcal{L}_{\mathrm{MSE}} &= \frac{1}{N}\sum_{i=1}^{N}
    \bigl\|\hat{\mathbf{y}}(t_i) - \mathbf{y}(t_i)\bigr\|^2,
    \label{eq:loss_mse} \\[4pt]
  \mathcal{L}_{\mathrm{smooth}} &= \frac{1}{N-1}\sum_{i=1}^{N-1}
    \left\|\frac{\hat{\mathbf{y}}(t_{i+1}) -
    \hat{\mathbf{y}}(t_i)}{\Delta t}\right\|^2,
    \label{eq:loss_smooth}
\end{align}


\noindent and $\lambda = 10^{-4}$.$\mathcal{L}_{\mathrm{MSE}}$ is the trajectory mean squared error and $\mathcal{L}_{\mathrm{smooth}}$ is a
smoothness penalty that encodes the physical prior that transformer temperatures cannot change rapidly, discouraging high-frequency oscillations in the predicted trajectory inconsistent with known thermal dynamics.The value $\lambda = 10^{-4}$ was set empirically; the most consequential sensitivity observed during development was to the ODE solver choice, fixed-step solvers such as \texttt{rk4} produced unstable integration trajectories, whereas the adaptive \texttt{dopri5} solver yielded stable training across all units.

Gradient stability is maintained through gradient clipping with maximum norm $1.0$. The optimizer is Adam with learning rate $\eta = 10^{-4}$, weight decay $10^{-5}$, and a step learning rate schedule with decay factor $\gamma = 0.9$ applied each epoch.

\subsubsection{Curriculum Learning}
\label{subsubsec:curriculum}


Training uses a sliding-window curriculum strategy in which the prediction horizon grows linearly from $W_{\min} = 5$ steps ($\approx 1.25\,\mathrm{h}$) at epoch~1 to $W_{\max} = 192$ steps ($= 2\,\mathrm{days}$) at the final epoch. The window bounds reflect known transformer thermal time constants: the minimum of 1.25\,h corresponds to the lower end of typical thermal response times, while the maximum of 2\,days captures full diurnal load cycles. Starting with short horizons and gradually increasing difficulty accelerates convergence by allowing the network to first learn short-range thermal dynamics before being exposed to the full integration length, following the curriculum learning principle of \cite{Bengio2009Curriculum}.

\subsection{LSTM Baseline}
\label{subsec:lstm_impl}

A three-layer stacked LSTM with hidden dimension~256 and dropout~0.1 is used as
the baseline. At each step $t$ the network receives
$[\mathbf{y}(t),\,\mathbf{x}(t)]$ as input and produces a prediction of
$\mathbf{y}(t+1)$, directly mirroring the input--output convention of the
Neural ODE vector field. This ensures the comparison is not confounded by
differences in the information available at each prediction step.

The training context window is fixed at 192~steps (2~days), equal to the Neural
ODE curriculum upper bound (Section~\ref{subsubsec:curriculum}), so both models
see identical maximum temporal context. Training uses scheduled
sampling with teacher-forcing ratio decaying
linearly from 1.0 at epoch~1 to 0.0 at epoch~60, producing a fully
autoregressive model by the final epoch and closing the train-to-evaluation
distribution gap caused by pure teacher forcing. All optimizer and
regularization settings (Adam, $\eta=10^{-4}$, weight decay $10^{-5}$,
step-LR $\gamma=0.9$, gradient clip norm~1.0, 60~epochs, batch size~1024) are
identical to those in Section~\ref{subsec:training}.

At evaluation time the LSTM is initialized with $\mathbf{y}(t_0)$ and a zero
hidden state, then rolled out autoregressively over the forecast window using
ground-truth exogenous inputs, imposing identical information constraints to
the Neural ODE.





\section{Results and Discussion}
\label{sec:results}

The central motivation of this work is that existing approaches to transformer
thermal modelling each carry fundamental limitations: high-fidelity numerical
methods such as FEM and CFD are computationally prohibitive and require complete
knowledge of internal geometry; lumped-parameter models depend on
transformer-specific constants that are rarely known a priori; and purely
data-driven models such as LSTMs learn statistical correlations from training
data, making them brittle outside the training distribution and prone to
physically inconsistent predictions over long horizons. The Neural ODE framework
proposed here is designed to occupy the gap between these extremes: it encodes
the continuous-time heat-transfer structure of
Equations~\eqref{eq:winding_ode}--\eqref{eq:oil_ode} directly into the
architecture, the network always outputs $d\mathbf{y}/dt$ and the temperature
trajectory is obtained by numerical integration. This structural inductive bias
is expected to yield smoother, more physically plausible long-horizon predictions
and better generalization across transformers with different cooling
configurations.

\subsection{Neural ODE Performance}
\label{subsec:node_performance}

\begin{figure}[h]
    \centering
    \includegraphics[width=\columnwidth]{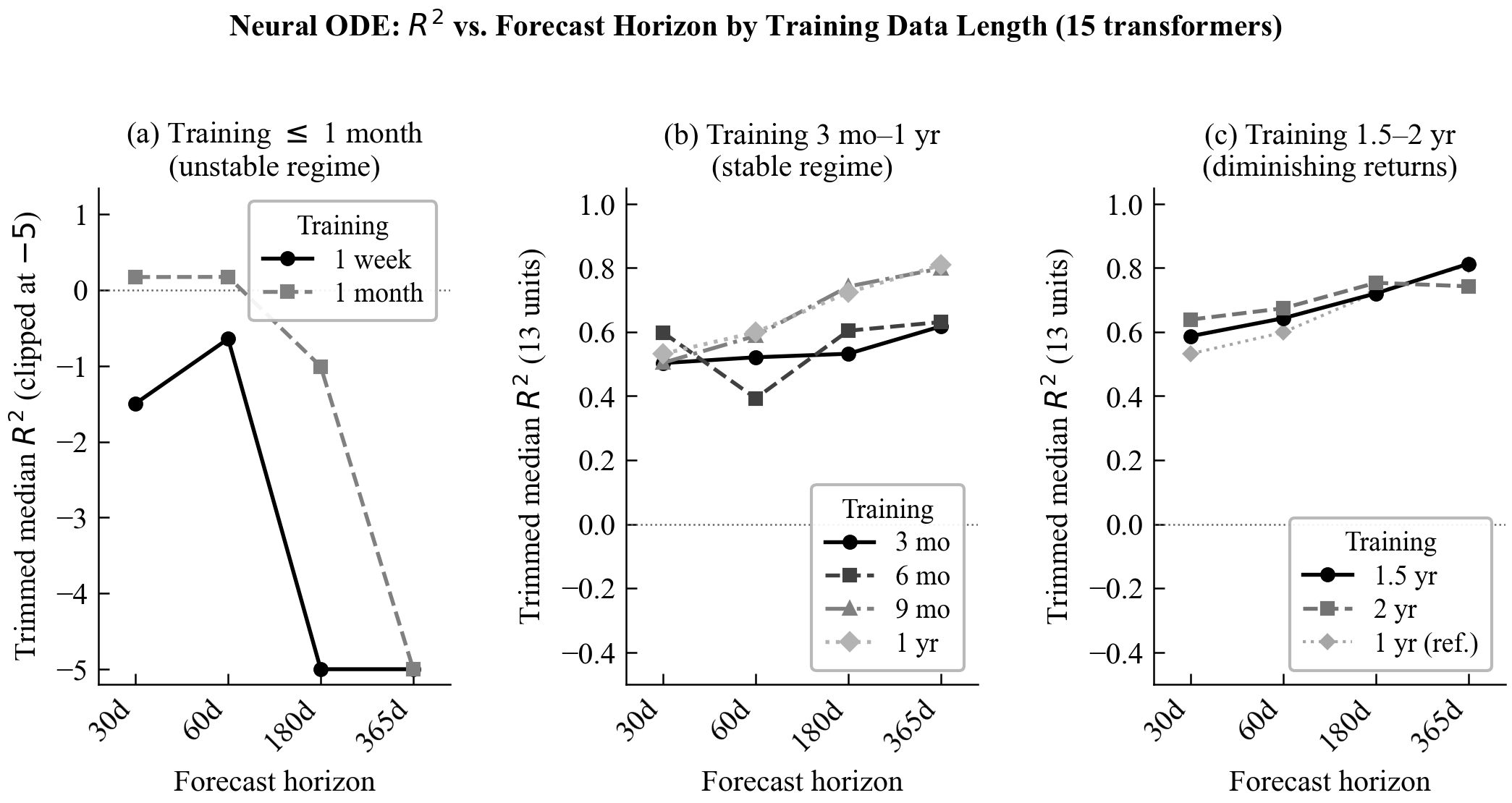}
    \caption{Neural ODE $R^2$ vs.\ forecast horizon by training data length, showing the trimmed median across all 15 transformer units.}
    \label{fig:r2_regimes}
\end{figure}

\begin{figure}[h]
    \centering
    \includegraphics[width=\columnwidth]{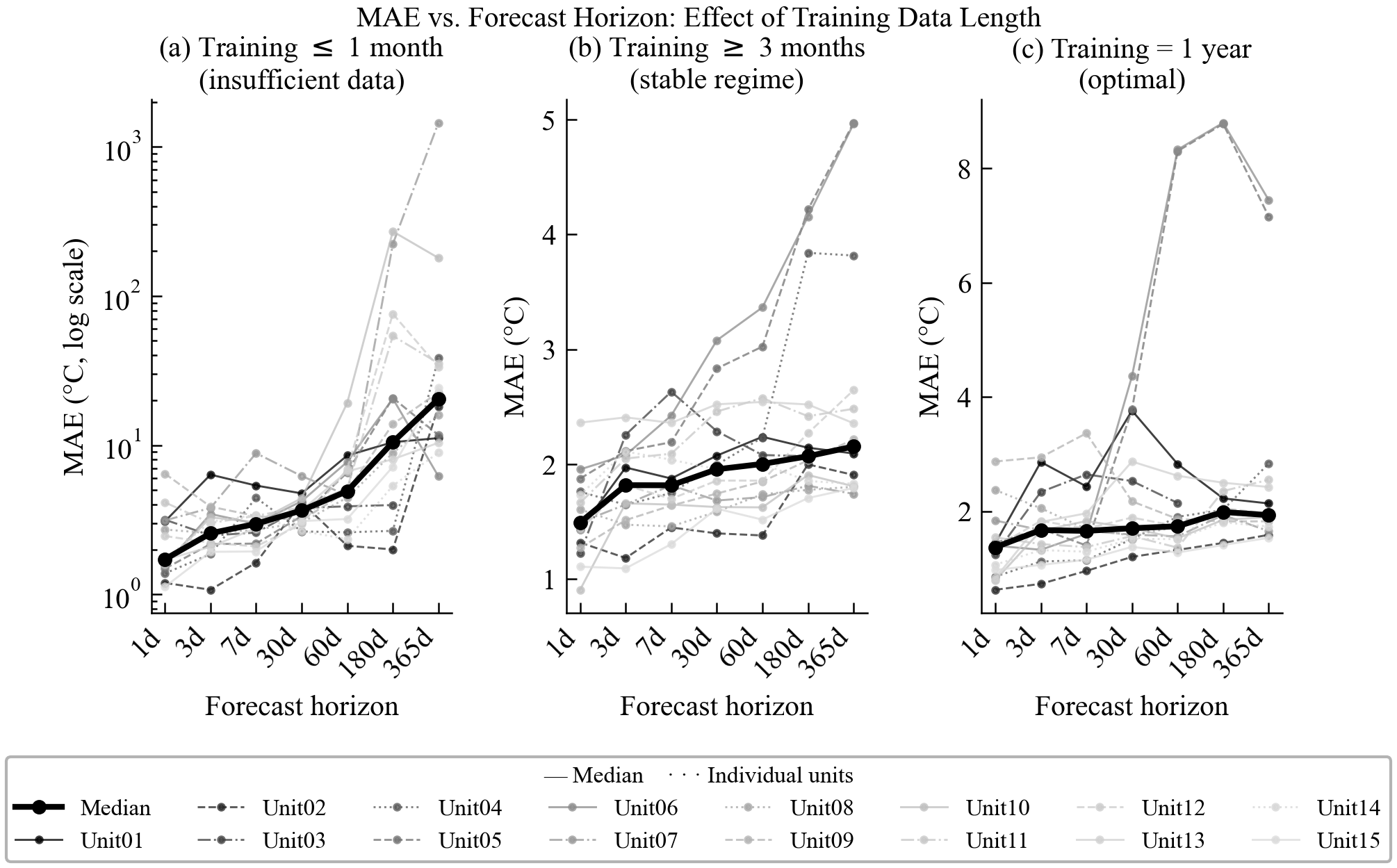}
    \caption{Neural ODE MAE (\textdegree C) vs.\ forecast horizon for all 15 units (grey)
             and their median (black). Panel~(a): training $\leq$1~month
             (log scale), (b): training $\geq$3~months, (c): training = 1~year.}
    \label{fig:mae_units}
\end{figure}

Figures~\ref{fig:r2_regimes} and~\ref{fig:mae_units} characterize Neural ODE
forecast quality as a function of training data length and forecast horizon.
Three distinct performance regimes emerge from the data.

In the unstable regime (training $\leq$ 1 month), with one week of training data
the median $R^2$ across all forecast horizons is $-3.29$, and 16 of the 105
evaluated (unit, forecast horizon) combinations produce $R^2 < -100$, indicating
catastrophic trajectory divergence (Figure~\ref{fig:r2_regimes}a). With one week of training data, corresponding to approximately 670 samples, 
the network parameters are far from convergence and the learned derivatives 
become numerically unstable over long integration horizons, producing 
catastrophic trajectory divergence. This is not a fundamental limitation 
of the framework but a minimum data requirement that is straightforward 
to enforce in deployment. The MAE is likewise unreliable, with median values
reaching $4.55$\,\textdegree C, more than three times the value achieved under
adequate training (Figure~\ref{fig:mae_units}a). At one month, median performance
improves to $R^2 = -0.75$, still well below the threshold of useful prediction.

From three months of training onward the Neural ODE enters a stable and useful
regime (Figure~\ref{fig:r2_regimes}b). Median $R^2$ reaches $+0.387$ at three
months and improves monotonically to $0.521$ at one year, while median MAE drops
from $2.07$\,\textdegree C to $1.73$\,\textdegree C over the same range
(Figure~\ref{fig:mae_units}b,c). To put these numbers in operational context:
forecasts are on average within approximately $2$\,\textdegree C of the true
temperature over an entire year of open-loop integration, which is physically
meaningful for asset management decisions such as load scheduling and insulation
ageing assessment, where temperature thresholds typically carry tolerances of
5--10\,\textdegree C. This threshold has a natural physical interpretation: three
months of data at 15-minute resolution amounts to approximately 8{,}640 samples
spanning a full seasonal quarter. Transformer thermal behavior is strongly coupled
to ambient temperature, which in Norway varies by up to $30$\,\textdegree C
between winter and summer. A model trained on less than a full season cannot have
observed the complete range of the ambient-to-oil temperature relationship and will
therefore generalize poorly to operating conditions it has never seen. 



Beyond one year, performance plateaus at approximately $R^2 = 0.59$ for forecast horizons of 30~days and longer, and median MAE stabilizes around $1.96$--$1.99$\,\textdegree C (Figure~\ref{fig:r2_regimes}c). This suggests the Neural ODE has extracted the available thermal dynamics within one year of data,
consistent with the model having observed all four seasons, and a one-year training window is therefore sufficient and recommended for deployment.

Across all training regimes, forecast quality degrades with increasing horizon as
errors accumulate in open-loop integration (Figure~\ref{fig:mae_units}). The degradation is slower for the Neural ODE than for the LSTM because the ODE structure produces inherently smooth, physically bounded trajectories. Short-horizon forecasts (1--7 days) remain challenging for both models with negative median $R^2$.
The Neural ODE is therefore best understood as a medium-to-long horizon virtual sensor (30--365 days).

\subsection{Comparison with LSTM Baseline}
\label{subsec:comparison}

\begin{figure}[h]
    \centering
    \includegraphics[width=\columnwidth]{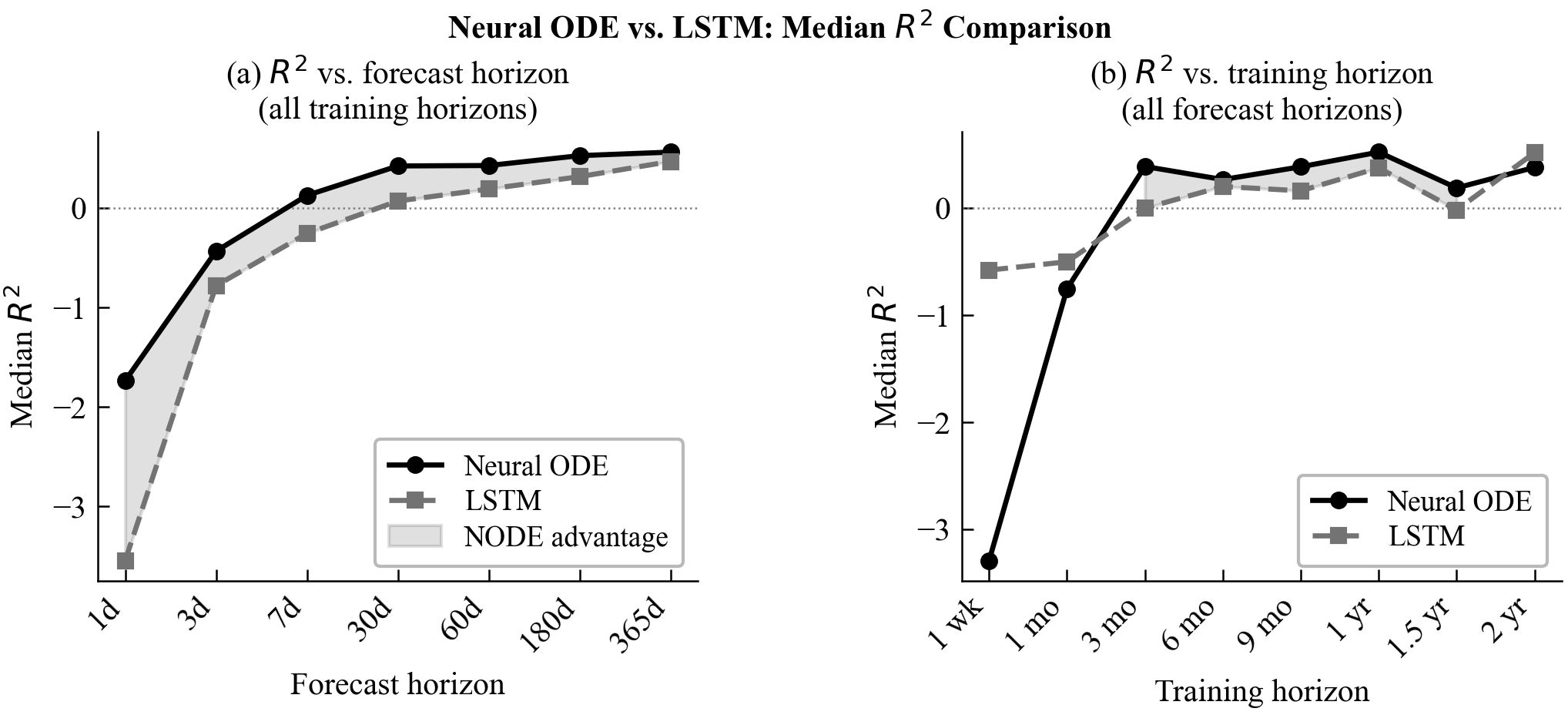}
    \caption{Neural ODE vs.\ LSTM $R^2$ comparison (15 transformers). Panel~(a):
         vs.\ forecast horizon (training $\geq$3~months). Panel~(b): vs.\
         training horizon at 365-day forecast. Shaded region marks Neural ODE
         advantage; downward arrows mark divergence (clipped at $-3$).}
    \label{fig:node_vs_lstm}
\end{figure}

Tables~\ref{tab:by_training}-- \ref{tab:by_forecast} and
Figure~\ref{fig:node_vs_lstm} provide a systematic head-to-head comparison
across all 823 evaluated combinations of unit, training horizon, and forecast horizon.
Bold entries in the tables indicate the best performing model for that row.

Overall, the $R^2$ indicate that both models struggle to provide useful models for less than 2 months of training data. At one week, LSTM median $R^2 = -0.58$, meaning the model explains less variance than a flat mean prediction. Hence we do not discuss best performance for these. At two years, LSTM median $R^2 = 0.52$, which is marginally higher than NODE, but both models perform comparably in absolute terms.

\rowcolors{2}{rowgray}{white}

\begin{table*}
\centering
\setlength{\tabcolsep}{12pt}
\caption{NODE vs.\ LSTM performance by training horizon (all 15 units, all 7 forecast horizons).
         Bold: better value. BP = best performance rate. N = NODE. L = LSTM.}
\begin{tabular}{lrrrrrr}
\toprule
\rowcolor{white}
\textbf{Training horizon} & \textbf{BP $R^2$} & \textbf{BP MAE} &
\textbf{Med $R^2$ N} & \textbf{Med $R^2$ L} &
\textbf{MAE N (\textdegree C)} & \textbf{MAE L (\textdegree C)} \\
\hline
1 week    & 34\% & 38\% & $-3.29$          & $\mathbf{-0.58}$ & 4.55          & \textbf{4.03} \\
1 month   & 61\% & 69\% & $-0.75$          & $\mathbf{-0.50}$ & \textbf{2.41} & 3.79 \\
3 months  & 84\% & 86\% & $\mathbf{+0.39}$ & $-0.00$          & \textbf{2.07} & 2.92 \\
6 months  & 65\% & 71\% & $\mathbf{+0.27}$ & $+0.20$          & \textbf{1.92} & 2.42 \\
9 months  & 78\% & 80\% & $\mathbf{+0.38}$ & $+0.16$          & \textbf{1.83} & 2.33 \\
1 year    & 64\% & 72\% & $\mathbf{+0.52}$ & $+0.37$          & \textbf{1.73} & 2.03 \\
1.5 years & 60\% & 58\% & $\mathbf{+0.19}$ & $-0.02$          & \textbf{1.99} & 2.04 \\
2 years   & 49\% & 44\% & $+0.38$          & $\mathbf{+0.52}$ & 1.96          & \textbf{1.81} \\
\bottomrule
\end{tabular}
\label{tab:by_training}
\end{table*}
\rowcolors{0}{}{}

Overall, across all 823 combinations, the Neural ODE achieves higher $R^2$ in 508 cases (61.7\%) and lower MAE in 533 cases (64.8\%). In the operationally relevant stable regime (training $\geq 3$ months, forecast $\geq 30$ days), these figures rise to 65.2\% on $R^2$ and 70.1\% on MAE, with median $R^2 = 0.591$ versus $0.493$ for the LSTM and median MAE $= 2.04$\,\textdegree C versus $2.45$\,\textdegree C -- a difference of $0.41$\,\textdegree C. On MAE, the Neural ODE is the better model at every forecast horizon without exception (Table~ \ref{tab:by_forecast}), even at 365 days where the $R^2$ best performance rate is marginally below 50\%.

\rowcolors{2}{rowgray}{white}
\begin{table*}
\centering
\setlength{\tabcolsep}{12pt}
\caption{NODE vs.\ LSTM performance by forecast horizon (all 15 units, all 8 training horizons).
         Bold: better MAE or $R^2$. BP = best performance rate. N = NODE. L = LSTM.}
\begin{tabular}{lrrrrrr}
\toprule
\rowcolor{white}
\textbf{Forecast horizon} & \textbf{BP $R^2$} & \textbf{BP MAE} &
\textbf{Med $R^2$ N} & \textbf{Med $R^2$ L} &
\textbf{MAE N (\textdegree C)} & \textbf{MAE L (\textdegree C)} \\
\hline
1 day    & 72\% & 73\% & $-1.74$ & $-3.54$ & \textbf{1.52} & 2.25 \\
3 days   & 61\% & 61\% & $-0.44$ & $-0.78$ & \textbf{1.82} & 2.19 \\
7 days   & 67\% & 67\% & $+0.13$ & $-0.26$ & \textbf{1.90} & 2.26 \\
30 days  & 68\% & 73\% & $+0.42$ & $+0.07$ & \textbf{2.09} & 2.52 \\
60 days  & 62\% & 68\% & $+0.43$ & $+0.19$ & \textbf{2.10} & 2.79 \\
180 days & 51\% & 57\% & $+0.52$ & $+0.31$ & \textbf{2.45} & 3.16 \\
365 days & 49\% & 52\% & $+0.56$ & $+0.47$ & \textbf{2.88} & 3.46 \\
\bottomrule
\end{tabular}

\label{tab:by_forecast}
\end{table*}
\rowcolors{0}{}{}

The four training regimes tell a clear story. At one week of training neither
model is usable: LSTM median $R^2 = -0.58$ and Neural ODE median $R^2 = -3.29$,
both well below zero -- neither exceeds a flat mean prediction. From three months
onward the Neural ODE clearly separates from the LSTM
(Figure~\ref{fig:node_vs_lstm}b): it already achieves median $R^2 = +0.39$ and
MAE $= 2.07$\,\textdegree C, while the LSTM median $R^2$ is effectively zero.
Three months spans a full seasonal quarter, giving the Neural ODE sufficient
exposure to the ambient--thermal coupling that governs transformer behavior; the
best performance rate reaches 84\% on $R^2$ and 86\% on MAE -- the highest of any training
horizon. At one year NODE median $R^2 = 0.52$ versus LSTM $= 0.37$
(Figure~\ref{fig:node_vs_lstm}a), and at two years the LSTM narrowly wins on
median $R^2$ (0.52 vs.\ 0.38) though the Neural ODE retains lower MAE throughout.

\subsection{Per-unit Analysis}
\label{subsec:per_unit}

Table~\ref{tab:node_lstm_per_unit} reports per-unit median performance in the stable regime (training horizon $\geq$ 3 months; forecast horizon $\geq$ 30 days) across all fifteen units. The Neural ODE attains higher median $R^2$ on 14 of 15 units and lower median MAE on all 15 units, indicating both stronger variance explanation and more accurate absolute predictions in the operationally relevant regime. The highest Neural ODE median $R^2$ values are obtained for U01, U09, and U15.

\rowcolors{2}{rowgray}{white}
\begin{table}[H]
\centering
\small
\caption{Per-unit performance comparison between NODE and LSTM in the stable regime (training horizon $\geq$ 3 months; forecast horizon $\geq$ 30 days). Higher $R^2$ and lower MAE indicate better performance. The better performance of each pair is shown in bold.}
\label{tab:node_lstm_per_unit}
\resizebox{\columnwidth}{!}{
\begin{tabular}{lcccc}
\toprule
\textbf{Unit} & \textbf{$R^2$ (NODE)} & \textbf{$R^2$ (LSTM)} & \textbf{MAE (NODE)} & \textbf{MAE (LSTM)} \\
\midrule
U01 & \textbf{0.909} & 0.899 & \textbf{2.09} & 2.34 \\
U02 & \textbf{0.717} & 0.713 & \textbf{1.54} & 1.80 \\
U03 & \textbf{0.561} & 0.519 & \textbf{2.13} & 2.41 \\
U04 & \textbf{0.607} & 0.477 & \textbf{2.40} & 3.37 \\
U05 & \textbf{0.214} & -0.092 & \textbf{3.71} & 4.92 \\
U06 & \textbf{0.159} & -0.148 & \textbf{4.20} & 5.78 \\
U07 & \textbf{0.561} & 0.433 & \textbf{1.73} & 2.00 \\
U08 & \textbf{0.732} & 0.668 & \textbf{1.76} & 1.94 \\
U09 & \textbf{0.825} & 0.692 & \textbf{1.95} & 2.49 \\
U10 & \textbf{0.519} & 0.397 & \textbf{1.75} & 1.87 \\
U11 & \textbf{0.157} & 0.101 & \textbf{2.45} & 3.02 \\
U12 & \textbf{0.530} & 0.504 & \textbf{2.04} & 2.22 \\
U13 & \textbf{0.466} & 0.004 & \textbf{2.48} & 3.20 \\
U14 & 0.475 & \textbf{0.699} & \textbf{1.90} & 1.94 \\
U15 & \textbf{0.773} & 0.694 & \textbf{1.68} & 1.85 \\
\bottomrule
\end{tabular}
}
\end{table}
\rowcolors{0}{}{}

Units~05 and~06 remain the most challenging cases for both models. Although the Neural ODE still attains positive median $R^2$ on both units, performance is weak relative to the rest of the power transfomers, and the LSTM yields negative median $R^2$ in both units. Aside from U14, where the LSTM achieves higher median $R^2$ ($0.699$ versus $0.475$), the Neural ODE outperforms the LSTM on median $R^2$ for every unit. Even on U14, the Neural ODE retains a slightly lower median MAE (1.90\,\textdegree C versus 1.94\,\textdegree C). These results indicate that the Neural ODE delivers the more consistent per-unit model overall.

\begin{figure}[!t]
    \centering
    \includegraphics[width=0.88\columnwidth]{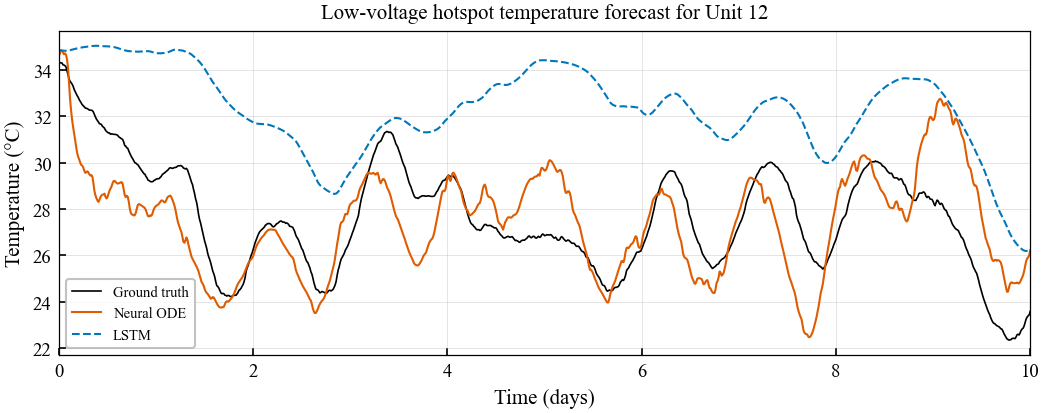}
    \caption{Zoomed-in comparison of predicted and measured low-voltage hotspot temperature for Unit~12 over a selected 10-day forecast interval.}
    \label{fig:unit12_lhs_zoom}
\end{figure}

Figure~\ref{fig:unit12_lhs_zoom} provides a zoomed-in qualitative comparison for Unit~12 low-voltage hotspot temperature over a selected 10-day forecast interval. 
For the shown interval, the Neural ODE follows the measured trajectory more closely, while the LSTM poorly performing.

\begin{figure*}[tp]
    \centering
    \includegraphics[width=\textwidth]{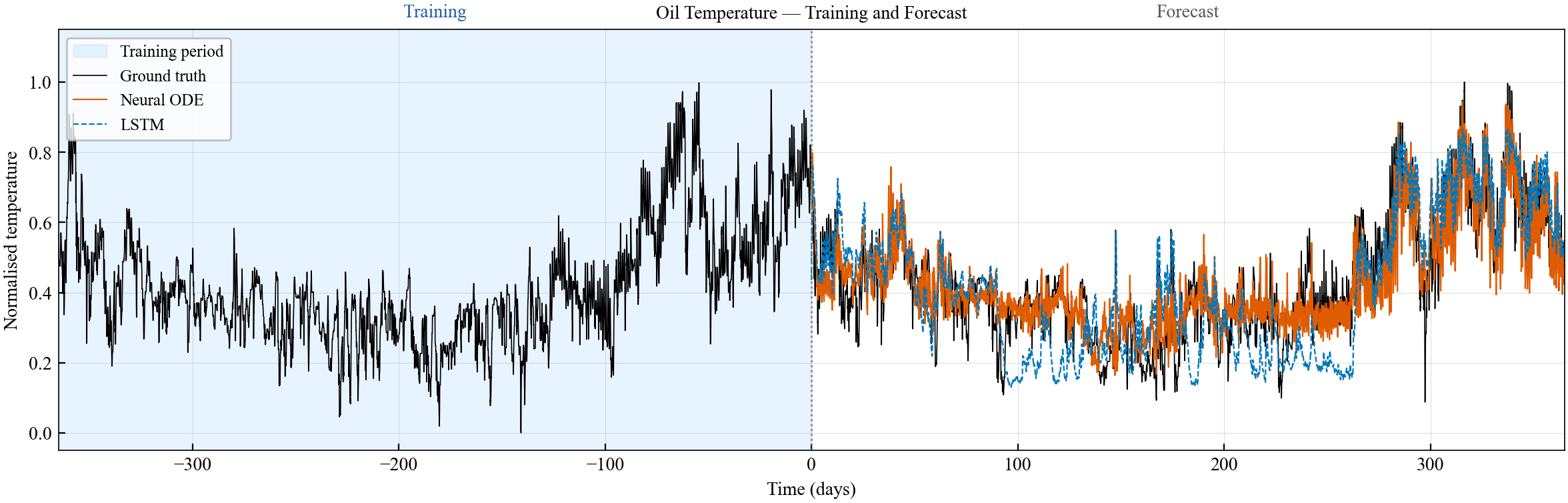}
    \caption{Oil temperature training and forecast for Unit~13 (1-year
             training, 365-day forecast). Training period shaded ($t < 0$),
             forecast period white ($t \geq 0$). Temperature normalised
             to $[0,1]$.}
    \label{fig:unit13_prediction}
\end{figure*}

Figure~\ref{fig:unit13_prediction} illustrates the qualitative difference
between the two models for Unit~13, which illustrates the Neural ODE advantage among units with positive $R^2$ for both models
($\Delta R^2 = +0.211$). Over the 365-day forecast, the Neural ODE tracks the thermal trend closely throughout, while the LSTM diverges progressively during the low-temperature period before partially recovering.
This is consistent with the broader finding that the Neural ODE's physical structure produces more stable long-horizon trajectories
compared to the purely data-driven LSTM baseline.

\section{Conclusion}
\label{sec:conclusion}

This paper presented a physics-aware Neural ODE framework for virtual
temperature sensing in power transformers. By encoding the continuous-time
heat-transfer structure of the winding, oil, ambient system directly into the
model architecture, the framework produces smooth, physically grounded thermal
trajectories and generalizes across fifteen heterogeneous transformer units. This enables deployment in transformer condition monitoring systems where direct hotspot sensing is unavailable. By providing reliable virtual temperature estimates, the framework can reduce the need for dense physical sensor deployment, lowering monitoring cost while still supporting early detection of abnormal thermal behaviour. This is particularly relevant for distribution-level assets, where large fleets of transformers must be monitored with limited instrumentation and maintenance resources.

As summarized in Table~\ref{tab:by_forecast} and Figure~\ref{fig:node_vs_lstm},
the Neural ODE consistently outperforms the LSTM baseline in the
operationally relevant regime. Across stable operating conditions (training
horizons $\geq 3$ months and forecast horizons of 30--365~days), it achieves
median $R^2 = 0.591$ and median MAE $= 2.04$\,\textdegree C, with best performance rates of
65.2\% on $R^2$ and 70.1\% on MAE. The most significant practical advantage is
data efficiency: the Neural ODE produces useful forecasts from as little as
three months of training data, whereas the LSTM median $R^2$ remains near zero
in this regime. This property is particularly valuable for new or recently
instrumented units where long historical records are unavailable.

Table~\ref{tab:node_lstm_per_unit} further shows that the Neural ODE provides
the more consistent per-unit model overall, achieving higher median $R^2$ on
14 of 15 units and lower median MAE on all 15 units. This indicates that the
benefit of the physics-aware formulation is not limited to aggregate trends,
but extends across a heterogeneous fleet of transformer units.

The main limitation is a minimum data requirement: below roughly three months
of training data, the learned dynamics become unstable and long-horizon
integration produces unreliable forecasts. Future work will investigate warmup
strategies and regularization methods to reduce this threshold, as well as
multi-unit transfer learning to further lower per-unit data requirements, as well of investigate probabilistic extensions and online adaptation.

\section*{Acknowledgment}
This publication has been funded by the SFI NorwAI, (Centre for Research-based Innovation, 309834). The authors gratefully acknowledge the financial support from the Research Council of Norway and the partners of the SFI NorwAI, in particular Statnett who shared their data.

\section*{Nomenclature}
\small
\setlength{\tabcolsep}{3pt}
\renewcommand{\arraystretch}{0.86}
\begin{tabular}{ll}
$T$ & temperature \\
$T_{\mathrm{oil}}$ & top-oil temperature \\
$T_{\mathrm{w}}$ & winding temperature \\
$T_{\mathrm{hw}}$, $T_{\mathrm{lw}}$ & high-/low-voltage winding temperature \\
$T_{\mathrm{hhs}}$, $T_{\mathrm{lhs}}$ & high-/low-voltage hotspot temperature \\
$T_{\mathrm{amb}}$ & ambient temperature \\
$P$ & active power load \\
$P_{\mathrm{hv}}$, $P_{\mathrm{lv}}$ & high-/low-voltage active power load \\
$t$ & time \\
$\Delta t$ & sampling interval, 900\,s \\
$\mathbf{y}(t)$ & internal thermal state vector \\
$\mathbf{x}(t)$ & exogenous control input vector \\
$C_{p,w}$, $C_{p,o}$ & winding/oil thermal capacitance \\
$h_1$, $h_2$ & heat-transfer coefficients \\
$L_E(P)$ & electrical losses as function of load \\
$F$ & learned thermal dynamics function \\
$\theta$ & trainable parameters \\
$\mathcal{NN}_{\theta}$ & neural network vector field \\
$H$ & hidden dimension \\
$\mathcal{L}$ & composite training loss \\
$\lambda$ & smoothness penalty weight \\
$\eta$, $\gamma$ & learning rate and decay factor \\
$W_{\min}$, $W_{\max}$ & curriculum window bounds \\
$R^2$ & coefficient of determination \\
MAE, MSE & mean absolute/squared error \\
\end{tabular}
\normalsize

\bibliographystyle{apacite}

@article{Karniadakis2021,
  author  = {Karniadakis, George Em and Kevrekidis, Ioannis G. and Lu, Lu and Perdikaris, Paris and Wang, Sifan and Yang, Liu},
  title   = {Physics-informed machine learning},
  journal = {Nature Reviews Physics},
  year    = {2021},
  volume  = {3},
  pages   = {422--440},
  doi     = {10.1038/s42254-021-00314-5}
}

@article{Huang2022PINNsReview,
  author  = {Huang, Bin and Wang, Jianhui},
  title   = {Applications of physics-informed neural networks in power systems --- a review},
  journal = {IEEE Transactions on Power Systems},
  year    = {2023},
  volume  = {38},
  number  = {1},
  pages   = {572--588},
  doi     = {10.1109/TPWRS.2022.3162473}
}

@article{Bragone2022PINNsThermal,
  author  = {Bragone, Federica and Morshuis, Peter and Laneryd, Tor and Luvisotto, Michele and Morozovska, Kateryna},
  title   = {Physics-informed neural networks for modelling power transformer thermal dynamics},
  journal = {Electric Power Systems Research},
  year    = {2022},
  volume  = {211},
  pages   = {108194},
  doi     = {10.1016/j.epsr.2022.108194}
}

@inproceedings{Gao2023PhysicsGCN,
  author    = {Gao, Xinbo and others},
  title     = {Physics-informed graph convolutional neural network for power system state estimation and optimal power flow},
  booktitle = {Proceedings of the IEEE Power \& Energy Society General Meeting},
  year      = {2023},
  doi       = {10.1109/PESGM52003.2023.10252870}
}

@inproceedings{Chen2018NeuralODE,
  author    = {Chen, Ricky T. Q. and Rubanova, Yulia and Bettencourt, Jesse and Duvenaud, David},
  title     = {Neural Ordinary Differential Equations},
  booktitle = {Advances in Neural Information Processing Systems (NeurIPS)},
  volume    = {31},
  pages     = {6572--6583},
  year      = {2018},
  publisher = {Curran Associates, Inc.},
  doi       = {10.48550/arXiv.1806.07366}
}

@inproceedings{Kidger2020NeuralCDE,
  author    = {Kidger, Patrick and Morrill, James and Foster, James and Lyons, Terry},
  title     = {Neural Controlled Differential Equations for Irregular Time Series},
  booktitle = {Advances in Neural Information Processing Systems (NeurIPS)},
  volume    = {33},
  year      = {2020},
  pages     = {6696--6707},
  publisher = {Curran Associates, Inc.},
  doi       = {10.48550/arXiv.2005.08926}
}

@article{Kidger2021NeuralDE,
  author  = {Kidger, Patrick},
  title   = {On Neural Differential Equations},
  journal = {arXiv preprint},
  year    = {2022},
  eprint  = {2202.02435},
  archivePrefix = {arXiv},
  doi     = {10.48550/arXiv.2202.02435}
}

@inproceedings{He2016DeepResidual,
  author    = {He, Kaiming and Zhang, Xiangyu and Ren, Shaoqing and Sun, Jian},
  title     = {Deep Residual Learning for Image Recognition},
  booktitle = {Proceedings of the IEEE Conference on Computer Vision and Pattern Recognition (CVPR)},
  year      = {2016},
  pages     = {770--778},
  doi       = {10.1109/CVPR.2016.90}
}

@article{Swift2001,
  author  = {Swift, G. and Molinski, T. S. and Lehn, W.},
  title   = {A fundamental approach to transformer thermal modeling},
  journal = {IEEE Transactions on Power Delivery},
  year    = {2001},
  volume  = {16},
  number  = {2},
  pages   = {171--175},
  doi     = {10.1109/61.915490}
}

@article{Lundgaard2004,
  author  = {Lundgaard, Lars E. and Hansen, Willy and Linhjell, Dag and Painter, Thomas J.},
  title   = {Aging of oil-impregnated paper in power transformers},
  journal = {IEEE Transactions on Power Delivery},
  year    = {2004},
  volume  = {19},
  number  = {1},
  pages   = {230--239},
  doi     = {10.1109/TPWRD.2003.820175}
}

@article{Susa2005,
  author  = {Susa, Dejan and Lehtonen, Matti and Nordman, Hasse},
  title   = {Dynamic thermal modelling of power transformers},
  journal = {IEEE Transactions on Power Delivery},
  year    = {2005},
  volume  = {20},
  number  = {1},
  pages   = {197--204},
  doi     = {10.1109/TPWRD.2004.835258}
}

@article{Piercy1994,
  author  = {Piercy, R. and McNutt, W. J. and Arseneau, R. and Ouellette, J.},
  title   = {Application of a new thermal model to transformer loading},
  journal = {IEEE Transactions on Power Delivery},
  year    = {1994},
  volume  = {9},
  number  = {1},
  pages   = {133--140},
  doi     = {10.1109/61.277684}
}

@article{Oommen2009TransformerDesign,
  author  = {Oommen, T. V. and Claiborne, C. C. and Mullen, E. T.},
  title   = {Vegetable oil-based dielectric coolants},
  journal = {IEEE Industry Applications Magazine},
  year    = {2009},
  volume  = {15},
  number  = {5},
  pages   = {16--25},
  doi     = {10.1109/MIAS.2009.933600}
}

@article{williams2024cooling,
  author  = {Williams, John and others},
  title   = {Cooling system configurations and thermal performance in power transformers: a review},
  journal = {Electric Power Systems Research},
  year    = {2024},
  volume  = {228},
  pages   = {110050},
  doi     = {10.1016/j.epsr.2024.110050}
}

@article{Yan2025LSTM,
  author  = {Yan, Chao and others},
  title   = {LSTM-based transformer hot-spot temperature forecasting incorporating solar radiation and oil viscosity effects},
  journal = {IEEE Transactions on Power Delivery},
  year    = {2025},
  doi     = {10.1109/TPWRD.2024.3510000}
}

@article{JuarezBalderas2020HotSpot,
  author  = {Ju{\'a}rez-Balderas, E. and others},
  title   = {Hot-spot temperature estimation in power transformers using artificial neural networks},
  journal = {IEEE Access},
  year    = {2020},
  volume  = {8},
  pages   = {195409--195419},
  doi     = {10.1109/ACCESS.2020.3033911}
}

@article{Vilaithong2007NeuralNetTopOil,
  author  = {Vilaithong, R. and Tenbohlen, S. and Stirl, T.},
  title   = {Prediction of top-oil temperature and loss of life of power transformers by using neural network},
  journal = {International Journal of Electrical Engineering Education},
  year    = {2007},
  volume  = {44},
  number  = {4},
  pages   = {323--334},
  doi     = {10.7227/IJEEE.44.4.5}
}

@article{Temboa2022DataDriven,
  author  = {Temboa, Emmanuel and others},
  title   = {Data-driven thermal models for power transformers: TiDE and temporal convolutional networks versus IEC 60076-7},
  journal = {Electric Power Systems Research},
  year    = {2022},
  volume  = {209},
  pages   = {107985},
  doi     = {10.1016/j.epsr.2022.107985}
}

@article{Wei2017HotspotForecasting,
  author  = {Wei, Hao and Wang, Zheng and others},
  title   = {Hot-spot temperature forecasting of power transformers based on machine learning},
  journal = {IEEE Transactions on Power Delivery},
  year    = {2017},
  doi     = {10.1109/TPWRD.2016.2634002}
}

@article{Tan2019UltraShortTerm,
  author  = {Tan, Min and others},
  title   = {Ultra-short-term temperature forecasting of power transformers using deep learning},
  journal = {International Journal of Electrical Power \& Energy Systems},
  year    = {2019},
  volume  = {113},
  pages   = {105910},
  doi     = {10.1016/j.ijepes.2019.105910}
}

@article{Rubasinghe2023CNNLSTM,
  author  = {Rubasinghe, Gayan and others},
  title   = {A CNN-LSTM approach for transformer thermal state prediction},
  journal = {IEEE Transactions on Power Delivery},
  year    = {2023},
  doi     = {10.1109/TPWRD.2023.3250000}
}

@inproceedings{Bengio2009Curriculum,
  author    = {Bengio, Yoshua and Louradour, J{\'e}r{\^o}me and Collobert, Ronan and Weston, Jason},
  title     = {Curriculum Learning},
  booktitle = {Proceedings of the 26th International Conference on Machine Learning (ICML)},
  year      = {2009},
  pages     = {41--48},
  doi       = {10.1145/1553374.1553380}
}
\PHMbibliography{ijphm}

\section*{Biographies}

\noindent\textbf{Berk Hadzhamolla} is a PhD research fellow at the University of Oslo, Norway. His research focuses on physics-informed machine learning for industrial systems, applications in epidemiological modeling, digital twins.

\noindent\textbf{Alexander Johannes Stasik} is a Senior Research Scientist for Analytics and AI at SINTEF Digital, Norway, as well as associate professor for data science at the Norwegian University for Life Science NMBU, Norway. His research focuses on physics-informed machine learning, probabilistic modeling and quantum computing for real world problems.

\noindent\textbf{Signe Riemer-S{\o}rensen} is a Senior Research Scientist and Research Manager for Analytics and AI at SINTEF Digital, Norway, as well as co-director for the Norwegian Center of AI for Decisions. Her research focuses on physics-informed machine learning for industrial applications. 

\end{document}